\documentclass[conference]{IEEEtran}
\IEEEoverridecommandlockouts
\usepackage{cite}
\usepackage{amsmath,amssymb,amsfonts}
\usepackage{algorithmic}
\usepackage{hyperref}
\usepackage{graphicx}
\usepackage{textcomp}
\usepackage{wrapfig}
\usepackage{xcolor}
\usepackage{subfig}
\usepackage{pifont}
\newcommand{\cmark}{\ding{51}}%
\newcommand{\xmark}{\ding{55}}%
\usepackage{booktabs} 
\def\BibTeX{{\rm B\kern-.05em{\sc i\kern-.025em b}\kern-.08em
    T\kern-.1667em\lower.7ex\hbox{E}\kern-.125emX}}

\begin{document}

\title{Deep Semi-supervised Learning\\ for Time Series Classification}

\author{\IEEEauthorblockN{Jann Goschenhofer\textsuperscript{1,2}, Rasmus Hvingelby\textsuperscript{2}, David Ruegamer\textsuperscript{1}, Janek Thomas\textsuperscript{1}, Moritz Wagner\textsuperscript{2}, Bernd Bischl\textsuperscript{1,2}}
\IEEEauthorblockA{\textit{LMU Munich, Munich, Germany}\textsuperscript{1}\\
\textit{Fraunhofer Institute for Integrated Circuits (IIS), Erlangen, Germany}\textsuperscript{2}\\
jann.goschenhofer@stat.uni-muenchen.de}
}

\maketitle

\begin{abstract}
While deep semi-supervised learning has gained much attention in computer vision, limited research exists on its applicability in the time series domain.
In this work, we investigate the transferability of state-of-the-art deep semi-supervised models from image to time series classification.
We discuss the necessary model adaptations, in particular an appropriate model backbone architecture and the use of tailored data augmentation strategies.
Based on these adaptations, we explore the potential of deep semi-supervised learning in the context of time series classification by evaluating our methods on large public time series classification problems with varying amounts of labeled samples.
We perform extensive comparisons under a decidedly realistic and appropriate evaluation scheme with a unified reimplementation of all algorithms considered, which is yet lacking in the field. 
We find that these transferred semi-supervised models show significant performance gains over strong supervised, semi-supervised and self-supervised alternatives, especially for scenarios with very few labeled samples. 
\end{abstract}

\begin{IEEEkeywords}
Semi-supervised Learning, Time Series Classification, Data Augmentation
\end{IEEEkeywords}

\section{Introduction}
\label{sec:intro}
Time series classification (TSC) spans many real-world applications in domains from healthcare \cite{Rajkomar2018} over cybersecurity \cite{Susto2018} to manufacturing \cite{Dau2019}. Several algorithms for TSC have been proposed over the years \cite{Bagnall2017} \cite{Fawaz2019}.

In many real-world scenarios, time series data can be collected easily, but acquiring labels for this data is costly. 
For instance, in disease monitoring, sensor data are collected with low effort but the labelling of this data requires time-consuming work by medical experts \cite{Goschenhofer2019}. 
Semi-supervised learning (SSL) addresses this by leveraging large amounts of unlabeled data in combination with a small amount of labeled data when training machine learning (ML) models. 

Especially in computer vision, the advances in deep neural networks and the promised label efficiency of SSL have lead to the introduction of several innovative approaches for image data \cite{VanEngelen2020}.
While there is much work on classical semi-supervised models for TSC, research on the use of neural network-based SSL algorithms for TSC is still limited.

This motivates our main research question that we approach holistically in this work: \textit{Can we transfer well established deep semi-supervised models from the image to the time series domain?}
More specifically, we answer this question for the most prominent state-of-the-art SSL approaches, by proposing adaptions for MixMatch \cite{Berthelot2019}, Virtual Adversarial Training \cite{Miyato2018}, the Mean Teacher \cite{Tarvainen2017} and the Ladder Net \cite{Rasmus2015}.
These include the modification of a suitable backbone architecture as well as adaptions of an appropriate data augmentation strategy to account for the domain transfer of these models.  
For demonstration of the efficacy of our proposed frameworks we adhere to best practices for realistic evaluation of semi-supervised models and provide a fair and reliable model comparison with a high degree of practicality \cite{Oliver2018}. 

\subsection{Related Work}
\label{chap:literature}

\paragraph{Time Series Classification}
Over the past years, a variety of methods has been developed for TSC. A detailed overview on classical ML methods that were specifically developed for TSC
\cite{Grabocka2014}, \cite{Bagnall2015}, \cite{Kate2016} is provided in \cite{Bagnall2017}. 
An alternative approach towards TSC consists in the extraction of statistical features from the raw time series as the basis for training any strong classifier for tabular data \cite{Christ2018}.   
Also in deep learning, specific methods for time series classification have been developed \cite{Wang2017}, \cite{Fawaz2020}, \cite{Bai2018}.
A comprehensive overview on these recent developments can be found in \cite{Fawaz2019}.

\paragraph{Semi-Supervised Learning}
There exists a plethora of different concepts that extract additional information from unlabeled data via semi-supervision.
These range from the extension of supervised ML methods such as the semi-supervised Support Vector Machine \cite{Vapnik1998} or semi-supervised Boosting \cite{Mallapragada2008} to inherently semi-supervised methods such as Label Propagation \cite{Zhu2002}, Manifold Regularization \cite{Belkin2006} or Co-Training \cite{Blum1998}.
\cite{Chapelle2009} provide a detailed overview on these semi-supervised approaches.
There is also growing research on deep semi-supervised learning, mainly driven by the computer vision community. 
A recent overview and taxonomy on these developments are provided by \cite{VanEngelen2020}.
Amongst these are graph-based methods such as Deep Label Propagation \cite{Iscen2019}, SNTG \cite{Luo2018} or the extension of pseudo-labelling for deep learning \cite{VanEngelen2020}.
Further, there is growing research on regularization-based approaches following the rationale of adding an additional unsupervised regularization loss term to the initial supervised loss. 
The Mean Teacher \cite{Tarvainen2017} and its predecessors, Temporal Ensembling and the $\Pi$-Model \cite{Laine2016}, employ a consistency loss over the unlabeled samples to reward similar predictions for differently augmented versions of the same unlabeled sample. 
To overcome one drawback of those methods, the need for domain-dependent data augmentation strategies, Virtual Adversarial Training (VAT) \cite{Miyato2018} adds small perturbations to the input data to create an auxiliary unsupervised training target. 
MixMatch \cite{Berthelot2019} in turn combines different regularization strategies in one common framework.
These regularization-based approaches yield state-of-the-art performance on image classification benchmarks.

\paragraph{SSL for TSC}
Different classical semi-supervised models have been developed for TSC.
In their foundational work, \cite{Wei2006} propose an approach that combines pseudo-labelling with a nearest-neighbor model for imbalanced, binary TSC tasks.
This cluster-then-label \cite{VanEngelen2020} rationale for labeled and unlabeled time series via custom distance metrics is also employed in approaches such as DTW-D \cite{Chen2013}, SUCCESS \cite{Marussy2013} or LCLC \cite{Nguyen2011}.
Graph-based label propagation \cite{Zhu2002} is combined with time-series-specific distance metrics by \cite{Xu2015} and \cite{Wang2019} introduced the shapelet-based SSSL.

\paragraph{Deep SSL for TSC}
There has been recent developments on neural net-based approaches. 
A customized version of the LadderNet \cite{Rasmus2015} based on the FCN architecture \cite{Wang2017} was applied by \cite{Zeng2017} on three multivariate human activity recognition (HAR) datasets.
They report relative gains of the semi-supervised model over the supervised baselines for small amounts of labeled samples. 
To the best of our knowledge, \cite{Zeng2017} are the first to evaluate SSL methods on large, multivariate TSC datasets.
A self-supervised approach, where the model is jointly trained on an auxiliary forecasting task over the whole dataset next to the initial supervised classification task on the labeled data only, was introduced by \cite{Jawed2020}.
They build upon the benchmark of \cite{Wang2019} on a subset of smaller, univariate TSC datasets from the UCR repository \cite{Dau2019} and report state-of-the-art performance compared to the majority of above methods as well as a customized variant of the $\Pi-$Model \cite{Laine2016} that works on time series problems. 
In alignment with \cite{Zeng2017}, they report particularly strong model performance for the deep supervised baseline FCN \cite{Wang2017} trained on few labeled samples only reporting it to outperform all above mentioned classical semi-supervised models. 
This deep learning baseline outperforms all of the classical semi-supervised models and almost always beats the $\Pi-$Model.
We include this approach as a self-supervised baseline in our experiments. 

\paragraph{Limitations}
All existing model comparisons for semi-supervised TSC, despite the work of \cite{Zeng2017}, are limited to univariate time series datasets with a maximal size of $1000$ training samples. 
In contrast to computer vision research on SSL \cite{VanEngelen2020}, these model comparisons are conducted for one fixed relative amount of labeled samples in the vast majority of experiments, making it hard to deduce general information for different data situations.
They also do not align with the guidelines established by \cite{Oliver2018} for SSL on image data and do not include repeated model runs to account for randomness in the selection of labeled samples.
Another issue is the lack of publicly accessible implementations of the classical approaches to semi-supervised TSC, making it impossible to validate against these approaches. 
This in turn leads to the problem that model comparisons with existing methods solely rely on values reported in former work for the same datasets with partially opaque dataset splits and unlabelling procedures.

Our \textbf{main contributions} can be summarized as follows: 
1) We propose four new deep SSL algorithms for TSC and describe tuning parameters and meaningful data augmentation strategies. 
2) We investigate the applicability of deep SSL in the domain of TSC and provide insights in which settings the proposed methods work well and how they compare to existing approaches. 
3) Through these experiments we are able to identify two out of our four proposed methods that notably improve over existing approaches.

\section{From Images to Time Series}
\label{chap:2}

\subsection{Problem Formulation}

We define an equidistant time series as $x^{(i)}=\{\{x_{1,1}^{(i)}, ..., x_{1,t}^{(i)}\}, ..., \{x_{c,1}^{(i)}, ..., x_{c,t}^{(i)}\}\}$, where $t$ describes the length and $c$ the amount of covariates such that $x^{(i)}\in \mathcal{X} \subseteq \mathbb{R}^{c \times t}$. 
For $c=1$ the time series is called univariate and for $c>1$ multivariate.
Next to the input space $\mathcal{X}$, we use $y^{(i)}\in \mathcal{Y}$ to denote a categorical variable in the target space $\mathcal{Y}$. 
The goal of SSL is to train a prediction model $f:\mathcal{X}\mapsto \mathcal{Y}$ on a dataset $\mathcal{D} = (\mathcal{D}^l, \mathcal{D}^{u})$ which consists of a labeled dataset $\mathcal{D}^l = \{(x^{(i)}, y^{(i)})\}_{i=1}^{n_l}$ and an unlabeled dataset $\mathcal{D}^{u} = \{x^{(i)}\}_{i=n_l+1}^{n}$ where $n=n_l+n_u$. 
We consider the case where $n_l \ll n_u$, as usual in SSL. 
Further, we define one batch of data as $\mathcal{B} \subset \mathcal{D}$, where $\mathcal{B}^l \subseteq \mathcal{D}^l$ contains the labeled samples and $\mathcal{B}^u \subseteq \mathcal{D}^u$ the unlabeled samples in that batch such that $\mathcal{B} = (\mathcal{B}^l, \mathcal{B}^u)$.

\subsection{Backbone Architecture} \label{chap:arch}

A basic building block in deep learning for images is a $3$-dimensional tensor,  whereas time series can be represented as $2$-dimensional tensors with channels corresponding to the number of covariates.
The extension of building blocks of powerful image classification architectures to TSC is thus straightforward, yet the right choice of a backbone architecture is crucial.
We propose the use of the Fully Convolutional Network (FCN)~\cite{Wang2017} as a backbone architecture as it was shown to outperform a variety of models on $44$ different TSC problems and is used in related work on semi-supervised TSC \cite{Jawed2020}. 
In all regularization-based semi-supervised methods discussed in Section~\ref{sec:methods}, except for the Ladder Net \cite{Rasmus2015}, the network architecture can be decoupled from the model training strategy.
This allows us to replace the backbone architecture of many of the established SSL methods from image classification with the FCN.
In case of the Ladder Net, we design the decoder as a mirrored version of the FCN encoder (see Section~\ref{sec:methods}).

\subsection{Data Augmentation}
\label{chap:daug}

One crucial component of regularization-based semi-supervised methods is the injection of random noise into the model. 
Data augmentation strategies $g(x^{(i)}), g: \mathcal{X} \mapsto \mathcal{X}$ should be designed such that they perturbate the input $x^{(i)}$ of a sample while preserving the meaning of its label $y^{(i)}$.
This can be achieved by utilizing inherent invariances in the data, e.g., rotations of images usually preserve the meaning of an image.
For images, invariances can be easily understood visually. 
In the time series domain, such invariances are not straightforward to understand, rendering the design of reasonable data augmentation strategies in this domain challenging. 
A set of data augmentation strategies for multivariate time series classification was introduced by \cite{Um2017} and evaluated on one HAR task. 
They show that the majority of strategies are beneficial, but some can deteriorate the model performance. 
To overcome the additional burden of choosing the right strategy, we propose the use of the RandAugment strategy \cite{Cubuk2020} which removes the need for a separate search phase. 
For each training batch, $N$ augmentation strategies are randomly chosen out of a set of $K$ possible policies. 
Next to $N$, a $magnitude$ hyperparameter is introduced which controls the augmentation intensity of the selected policies. 
We use the following set of augmentation policies \cite{Um2017}: warping in the time dimension, warping the magnitude, addition of Gaussian Noise and random rescaling.
We use RandAugment in this context following the rationale that even if a augmentation strategy is (not) label preserving, training with RandAugment with $N=1$ will still produce correct model updates in at least $\frac{K-1}{K}$ of the forward passes.
Early experiments in a fully supervised setting showed that the application of this data augmentation strategy improves model performance across all datasets used in our experiments. 

\subsection{Methods} \label{sec:methods}

The Mean Teacher \cite{Tarvainen2017} is the successor of a series of consistency-regularization-based models such as Temporal Ensembling or the $\Pi$-Model \cite{Laine2016} for SSL and was empirically shown to outperform its predecessors\cite{Oliver2018}. 
Thereby, a teacher model, that is an average of the consecutive student models, is used to enforce consistency in model predictions over the course of model training. 

Virtual Adversarial Training (VAT) \cite{Miyato2018} also focuses on consistency regularization. 
Similar to adversarial examples \cite{Goodfellow2014}, a small data perturbation is learned such that its addition to the initial data point is expected to yield the maximum change in the model's prediction.  
These perturbed model predictions are used as auxiliary labels for the unlabeled samples within a regularization term to enable model training on the whole data set. 
This approach is particularly interesting for the time series domain where visual inspection of the appropriateness of data augmentation policies is difficult, as it does not rely on data augmentation techniques. 

In MixMatch, various semi-supervised techniques such as data augmentation for consistency regularization, Mixup training \cite{Zhang2017} and pseudo-labeling are combined within one holistic approach \cite{Berthelot2019}.
It was empirically shown to perform well on image data, motivating our use of it in this work \cite{Berthelot2019}.

The Ladder Net by \cite{Rasmus2015} is a reconstruction-based SSL model and is inspired by denoising autoencoders \cite{Vincent2010}. 
In its core, it extends a supervised encoder model with a corresponding decoder network which allows for the calculation of an unsupervised reconstruction loss over the unlabeled samples enabling training on the whole dataset. 
The Ladder Net was previously extended to TSC problems \cite{Zeng2017} and is thus also part of this study. 

\section{Experimental Design}

\subsection{Baseline Models}

Next to shapelet- and distance-based methods \cite{Bagnall2017}, fitting standard ML methods on hand-crafted statistical features has been a widely used approach for TSC before the introduction of specific deep learning architectures for TSC \cite{Wang2017} \cite{Fawaz2020}.
We include a Random Forest and a Logistic Regression trained on features, extracted via the tsfresh framework \cite{Christ2018} from the time series, as baselines. \\
In addition, we train the FCN architecture \cite{Wang2017} on the labeled samples $\mathcal{D}^l$ based on the cross entropy loss as a supervised deep learning baseline model for our experiments.
To ensure a fair and reliable model comparison, we explicitly use the same architecture of this supervised baseline model as the backbone for all SSL approaches.
We also use the performance of a supervised FCN trained on the fully labeled datasets as an estimated upper bound for the model performance.\\
Furthermore, we evaluate the performance of the self-supervised approach that was recently introduced for TSC by \cite{Jawed2020}.
Thereby, an auxiliary forecasting task from the time series data $\mathcal{D}$ is created and combined with the initial classification task as a surrogate supervision signal allowing the use of unlabeled data in model training.
The model is then jointly trained on both tasks simultaneously.
Next to its re-implementation, we further extend their approach for multivariate TSC by increasing the amount of neurons in the surrogate model head accordingly. 
The direct comparison with this self-supervised approach is of special interest as it was shown to outperform classical semi-supervised approaches in a set of experiments on smaller TSC datasets \cite{Jawed2020}.

\subsection{Data Sets}

We evaluate the performance of the above described semi-supervised models on $6$ publicly available datasets. 
In contrast to previous work \cite{Jawed2020, Xu2015, Wang2019}, we explicitly focus on large datasets with at least $1000$ observations.
Their main characteristics are described in Table \ref{tab:data}.

\begin{table}[ht]
    \caption{Characteristics of the used data sets where \textit{c} refers to the amount of covariates, \textit{Size} to the size of the whole training data set and \textit{Length} to the length of the time series.}
    \centering
    \begin{tabular}{lrrrcc}
    \hline
     \textbf{Name} & \textbf{Classes} & \textbf{Size} & \textbf{Length} & \textbf{c} & \textbf{Balanced} \\
     \hline
     Crop            & 24 & 7,200 & 46 & 1 & \cmark \\
     ElectricDevices & 7 & 8,926 & 96 & 1 & \xmark \\
     FordB           & 2 & 3,636 & 500 & 1 & \cmark\\
     Pamap2          & 13 & 11,313 & 100 & 6 & \xmark\\
     WISDM           & 6 & 10,727 & 80 & 3 & \xmark\\
     Balanced SITS   & 6 & 35,064 & 46 & 1 & \cmark\\   
    \end{tabular}
    \vspace{0.3cm}
    \label{tab:data}
\end{table}

With Crop, ElectricDevices and FordB we include three of the largest datasets from the UCR Time Series Classification Repository \cite{Dau2019}. 
In addition, we use the two multivariate HAR datasets Pamap2 \cite{Reiss2012} and WISDM \cite{Kwapisz2011}. 
We also evaluate the models on a class-balanced version of the Satellite Image Time Series (SITS) dataset \cite{Petitjean2012}.

\begin{figure*}[!h]
    \centering
    \vspace{-0.5cm}
    \includegraphics[width=0.9\textwidth]{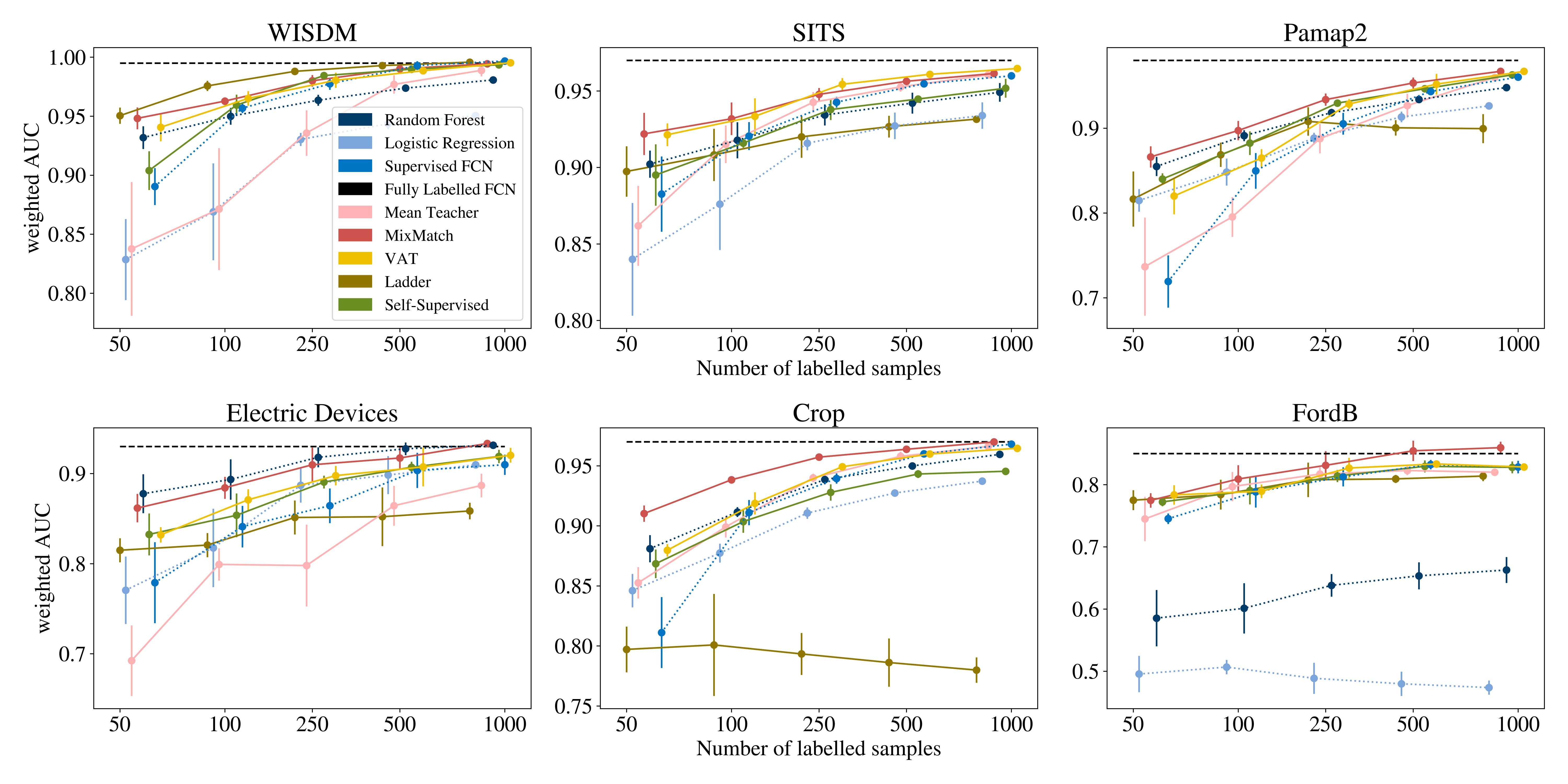}
    \caption{Performance of all models on the $6$ different datasets over various $n_l$ as presented in Table~\ref{tab:performance} in the appendix. 
    The horizontal line marks the performance of the fully labeled baseline, i.e. the supervised FCN model trained on the fully labeled dataset. 
    Dots represent the mean  wAUC and the vertical lines the standard deviation over $5$ repeated \textit{unlabeling} steps. 
    The performance of the baseline models are depicted as dotted, those of the semi-supervised models as solid lines. 
    Semi-supervised models clearly outperform the baseline models in settings with few labeled samples $n_l \in \{50, 100\}$ on all but the Electric Devices dataset.}
    \label{fig:main_results}
\end{figure*}

\subsection{Evaluation, Tuning and Implementation}
\label{chap:eval}
Due to special factors, such as the selection of the labeled data points, an unbiased and fair model comparison is particularly crucial to get a realistic perspective on the performance of the semi-supervised models \cite{VanEngelen2020}. 
We adhere to the guidelines for realistic evaluation of semi-supervised models by \cite{Oliver2018} to guarantee reliable and fair experimental results.
For performance evaluation of SSL models, the standard procedure is to split a fully labeled dataset $\mathcal{D}$ into labeled and unlabeled datasets $\mathcal{D}^l$ and $\mathcal{D}^u$ via artificial \textit{unlabeling} of $n_u$ randomly drawn samples \cite{VanEngelen2020}.
This way, semi-supervised data settings for different amounts of labeled samples $l$ are simulated. 
We unlabel in a stratified manner to retain the datasets' label distributions. 
For the following experiments, we split the evaluation of one model $f$ on one data set $D$ in two distinct phases.
\paragraph{Tuning Phase}
In the tuning phase, we tuned the model $f$ with one fixed amount of labeled samples to yield an optimal set of hyperparameters $\theta^*$. 
Thereby, $f$ was trained on a training dataset $\mathcal{D}_{train} = (\mathcal{D}_{train}^l, \mathcal{D}_{train}^u)$, where we fixed $|\mathcal{D}_{train}^l|=500$, and validated on a labeled holdout validation set $\mathcal{D}_{val}$. 
The choice of the size of $\mathcal{D}_{val}$ is subject to recent discussions \cite{Rasmus2015}, \cite{Oliver2018}, \cite{Zhai2019}.
Large $\mathcal{D}_{val}$ are expected to yield stable results for model tuning, which is important for many hyperparameter-sensitive semi-supervised models, but stands in contrast to the promised practicality of these models in settings with few labeled data.
First insights on this trade-off are are given by \cite{Oliver2018} and \cite{Zhai2019}, which empirically show in smaller experiments $|\mathcal{D}_{val}|= 1000$ to be a validation set size where variance in the performance estimates is still low enough to allow for reasonable model selection.
Following this, we set the size of the labeled validation set to $|\mathcal{D}_{val}|=1000$ which is rather small compared to recent literature where $|\mathcal{D}_{val}|\geq 4000$ \cite{Miyato2018}, \cite{Laine2016}, \cite{Tarvainen2017}.
A separate labeled test set $\mathcal{D}_{test}$ with $|\mathcal{D}_{test}|=2000$ is kept aside for the evaluation phase. 
Hyperband \cite{Li2016} with random sampling as implemented in the Optuna framework \cite{Akiba2019} was used for tuning, with a fixed budget of $100$ GPU hours for each deep learning model and dataset.
We measure model performance in terms of weighted Area under the Curve (wAUC) to account for model calibration and class imbalance.

\paragraph{Evaluation Phase}
In the evaluation phase, we train $f(\theta^*)$ on $\mathcal{D}_{train}$ with varying amounts of $n_l \in \{50, 100, 250, 500, 1000\}$ for a maximum of $25000$ model update steps, assuming $\theta^*$ is also a suitable hyperparameter set for amounts of labels $n_l \neq 500$ on which the model was not specifically tuned.
This evaluation scheme is in line with previous work on SSL for image data \cite{Berthelot2019}, \cite{Oliver2018}.
Model performance is tracked on $\mathcal{D}_{val}$ and the model checkpoint with the best validation performance is used for inference on the holdout $\mathcal{D}_{test}$. 
The selection of especially (un-)informative labeled samples can have a major effect on the model performance, especially for small $n_l$.
To account for potentially (un-)lucky selection of $\mathcal{D}_{train}^l$ in the \textit{unlabelling} split of $\mathcal{D}_{train} = (\mathcal{D}_{train}^l, \mathcal{D}_{train}^u)$, we repeat this \textit{unlabelling} step $5$ times.
In case of the ML baseline models, we use a Random Search with a budget of $100$ model evaluations for the tuning phase and evaluate them on the same set of values for $n_l$ in the evaluation phase.
See Table~\ref{tab:hyperparameters} in the appendix for the specific ranges. 
All deep learning models were implemented in a unified codebase\footnote{https://github.com/Goschjann/ssltsc} and trained using the Adam optimizer \cite{Kingma2014} with all parameters set to default values except the learning rate and weight decay. 
We implemented all deep learning models from scratch in one unified framework and validated our implementations based on performance metrics reported on image classification tasks..

\section{Experimental Results}
\label{sec:results}

Experimental findings are visualized in Figure~\ref{fig:main_results} and Table~\ref{tab:performance} in the appendix. 
The ranking of the various models for different $n_l$, averaged over the datasets, is shown in Figure~\ref{fig:ranking} and Table~\ref{tab:ranking} in the appendix. 

\paragraph{Semi-supervised models outperform supervised baselines}
Overall, our results show that semi-supervised models outperform baseline models especially for small amounts of labeled data.
This relative performance gain of semi-supervised over supervised models is decreasing with an increase in $n_l$ and we find that all models benefit from more labeled samples in most cases. 
This is in line with literature on SSL \cite{VanEngelen2020}.

\paragraph{Deep SSL translates well to TSC}
Following our experimental results in Figure~\ref{fig:main_results}, we deduct that \textit{transferring well-established semi-supervised models from the image to the time series domain is indeed possible.} 
We find that the deep semi-supervised models, especially the transferred MixMatch and VAT, show impressive performance gains over the deep supervised baseline model over all datasets up to $n_l =500$, even reaching the performance of the fully labeled baseline in few cases.
For instance, the Mixmatch model exceeds the deep supervised baseline by $0.16$ wAUC on the Pamap2 and by $0.10$ wAUC on the Crop dataset for $n_l=50$.
These findings again encourage our proposed transfer.

\begin{figure}
    \centering
    \vspace{-0.5cm}
    \includegraphics[width=0.35\textwidth]{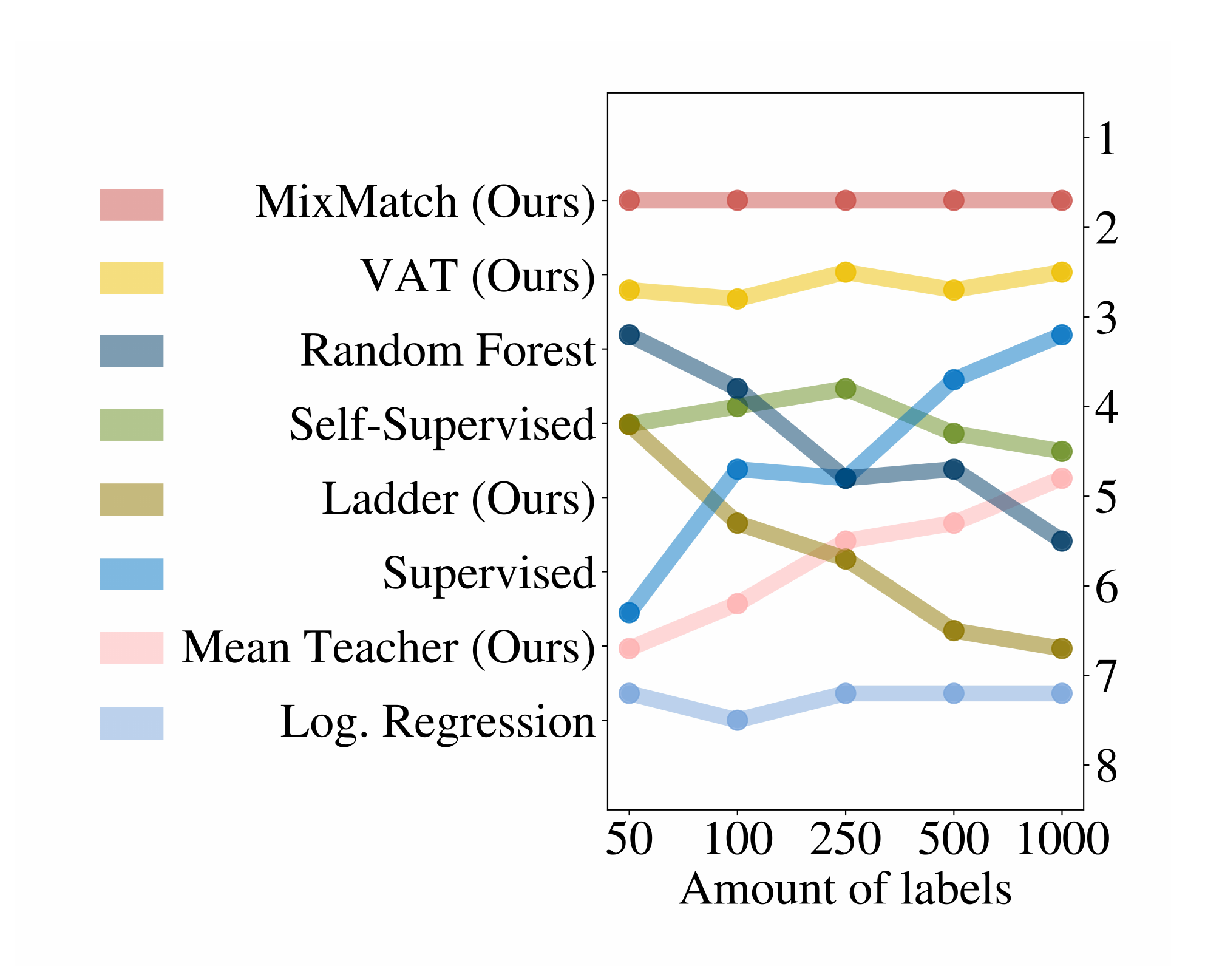}
    \vspace{-0.25cm}
    \caption{Average ranks of all models based on the wAUC over the $6$ datasets for varying $n_l$. Models are sorted by their strongest performance on $n_l=50$ and plotted with decreasing rank as indicated on the right vertical axis.}
    \label{fig:ranking}
\end{figure}

\paragraph{Strong baselines are crucial}
We find the use of strong baselines crucial for a realistic perspective on semi-supervised learning performance.
For instance, the Mean Teacher shows weak performance on the majority of datasets, often performing even worse than the supervised baseline. 
This is in line with results of \cite{Jawed2020}.
The strong performance of the Random Forest for small $n_l$ on the other hand also stresses the need for realistically strong supervised baselines.

\paragraph{Proposed methods outperform existing semi-supervised approaches}
While our results on the Ladder Net outperforming other supervised methods align with those of \cite{Zeng2017}, we also observe that the Ladder Net is notably worse compared to alternative SSL  algorithms we propose. 
This varying performance might be grounded in the large amount of hyperparameters of the Ladder Net and its sensitivity to different settings of those.

\paragraph{Proposed methods outperform self-supervised modeling}
Similar to \cite{Jawed2020}, we find their self-supervised approach to perform better or at least equally well compared to the deep supervised baseline model. 
Additionally, we are able to show that our extension towards multivariate time series also works well on the two multivariate datasets, WISDM and Pamap2.
The proposed approaches MixMatch and VAT furthermore consistently outperform this self-supervised approach across different amounts of labels on all $6$ datasets. 

\paragraph{Ranking of model performance similar to image domain}
In terms of model performance ranking, literature suggests that MixMatch performs better than VAT which again outperforms the Mean Teacher and the Ladder Net \cite{Berthelot2019}, \cite{Oliver2018}.
When ranking the algorithms across the datasets in Figure~\ref{fig:ranking}, we confirm this ranking in the TSC setting.

Our results show that the promised label efficiency of modern, deep semi-supervised model approaches translates well to TSC problems. 
Furthermore, these findings suggest the use of strong semi-supervised models from the image domain as these transferred models show stronger performance than the currently existing semi- and self-supervised approaches tailored towards TSC.
We believe that this work, also thanks to a strong focus on a fair and reliable model comparison, can serve as the basis for future research advances in semi-supervised learning for time series classification.

\section*{Acknowledgements}

This work was supported by the Bavarian Ministry of Economic Affairs, Regional Development and Energy through the Center for Analytics – Data – Applications (ADA-Center) within the framework of BAYERN DIGITAL II (20-3410-2-9-8) as well as the German Federal Ministry of Education and Research (BMBF) under Grant No. 01IS18036A.

\bibliographystyle{ieeetr}
\bibliography{literature}

\onecolumn
\section*{Appendix}

\subsection{Model Performance}

\begin{table}[h!]
\caption{Results for models over datasets with varying numbers of labels $n_l$. Performance is measured as weighted AUC. The best results for each $n_l$-dataset-combination are emphasized in bold with standard deviations over $5$ replications in brackets.}
\vspace{.5cm}
\resizebox{\columnwidth}{!}{

\begin{tabular}{l | llllll | llllll}
\toprule
Number of labels & \multicolumn{6}{l}{50} & \multicolumn{6}{l}{100} \\
Dataset &           Crop & Electric Devices &          FordB &         Pamap2 &           SITS &          WISDM &           Crop & Electric Devices &          FordB &         Pamap2 &           SITS &          WISDM \\
\hline
Model              &                &                 &                &                &                &                &                &                 &                &                &                &                \\
\midrule

Ladder             &  0.797 (0.019) &   0.815 (0.013) &  0.775 (0.016) &  0.816 (0.033) &  0.897 (0.017) &   0.95 (0.007) &  0.801 (0.042) &   0.821 (0.013) &  0.784 (0.024) &  0.869 (0.014) &  0.908 (0.017) &  0.976 (0.004) \\
Logistic Regression &  0.846 (0.014) &   0.771 (0.037) &  0.496 (0.029) &  0.815 (0.013) &   0.84 (0.037) &  0.829 (0.034) &  0.877 (0.008) &   0.817 (0.043) &  0.507 (0.012) &  0.848 (0.016) &   0.876 (0.03) &  0.869 (0.041) \\
Mean Teacher        &  0.853 (0.013) &   0.692 (0.039) &  0.745 (0.036) &  0.737 (0.058) &  0.862 (0.026) &  0.838 (0.057) &  0.899 (0.009) &   0.799 (0.018) &  0.797 (0.023) &  0.795 (0.023) &  0.915 (0.012) &  0.871 (0.052) \\
MixMatch           &   \textbf{0.910 (0.007)} &   0.862 (0.016) &  0.775 (0.012) &  \textbf{0.866 (0.013)} &  \textbf{0.922 (0.014)} &  \textbf{0.948 (0.009)} &  \textbf{0.938 (0.003)} &   0.884 (0.012) &  \textbf{0.809 (0.022)} &  \textbf{0.897 (0.011)} &  0.932 (0.011) &  0.963 (0.003) \\
Random Forest       &  0.881 (0.011) &   \textbf{0.878 (0.022)} &  0.585 (0.045) &  0.855 (0.011) &  0.902 (0.009) &   0.932 (0.01) &  0.911 (0.004) &   \textbf{0.893 (0.022)} &   0.601 (0.04) &  0.891 (0.006) &  0.918 (0.012) &   0.95 (0.007) \\
Self-Supervised     &  0.868 (0.012) &   0.832 (0.023) &  0.772 (0.007) &   0.84 (0.007) &   0.895 (0.02) &  0.904 (0.016) &  0.904 (0.009) &   0.854 (0.024) &   0.79 (0.022) &  0.882 (0.014) &  0.916 (0.003) &  0.959 (0.009) \\
Supervised         &   0.811 (0.03) &   0.779 (0.045) &  0.745 (0.009) &  0.719 (0.031) &  0.883 (0.025) &   0.89 (0.016) &   0.911 (0.01) &   0.841 (0.023) &  0.788 (0.025) &   0.85 (0.021) &  0.921 (0.009) &  0.957 (0.004) \\
VAT                &   0.88 (0.005) &   0.832 (0.009) &  \textbf{0.783 (0.016)} &   0.82 (0.022) &  0.921 (0.007) &  0.941 (0.012) &  0.919 (0.009) &   0.871 (0.012) &  0.789 (0.011) &   0.865 (0.01) &  \textbf{0.933 (0.012)} &  \textbf{0.965 (0.006)} \\

\bottomrule
\end{tabular}
}

\vspace{.5cm}

\resizebox{\columnwidth}{!}{
\begin{tabular}{l | llllll | llllll}
\toprule
Number of labels & \multicolumn{6}{l}{250} & \multicolumn{6}{l}{500} \\
Dataset &           Crop & Electric Devices &          FordB &         Pamap2 &           SITS &          WISDM &           Crop & Electric Devices &          FordB &         Pamap2 &           SITS &          WISDM \\
\hline
Model              &                &                 &                &                &                &                &                &                 &                &                &                &                \\
\midrule

Ladder             &  0.793 (0.017) &   0.851 (0.019) &  0.808 (0.028) &  0.908 (0.017) &   0.92 (0.014) &  0.988 (0.001) &   0.786 (0.02) &   0.852 (0.033) &  0.809 (0.007) &  0.901 (0.009) &  0.927 (0.007) &  0.993 (0.001) \\
Logistic Regression &  0.911 (0.005) &   0.887 (0.019) &  0.489 (0.025) &  0.888 (0.006) &  0.916 (0.005) &   0.93 (0.006) &  0.927 (0.002) &   0.898 (0.021) &   0.48 (0.019) &  0.913 (0.006) &  0.927 (0.009) &  0.944 (0.004) \\
Mean Teacher        &   0.94 (0.004) &   0.798 (0.045) &  0.817 (0.011) &  0.888 (0.017) &  0.943 (0.007) &  0.936 (0.019) &  0.958 (0.004) &   0.864 (0.022) &  0.823 (0.007) &  0.927 (0.014) &  0.953 (0.003) &  0.977 (0.008) \\
MixMatch           &  \textbf{0.957 (0.003)} &    0.910 (0.019) &  \textbf{0.831 (0.023)} &  \textbf{0.934 (0.007)} &  \textbf{0.948 (0.004)} &   0.980 (0.005) &  \textbf{0.964 (0.003)} &   0.917 (0.013) &  \textbf{0.854 (0.016)} &  \textbf{0.953 (0.006)} &  0.956 (0.002) &   0.990 (0.003) \\
Random Forest       &  0.939 (0.004) &   \textbf{0.918 (0.011)} &  0.638 (0.018) &  0.918 (0.005) &  0.934 (0.007) &  0.963 (0.005) &   0.950 (0.003) &   \textbf{0.928 (0.007)} &  0.653 (0.022) &  0.934 (0.004) &  0.942 (0.007) &  0.974 (0.001) \\
Self-Supervised     &  0.928 (0.007) &   0.891 (0.007) &   0.814 (0.01) &   0.930 (0.001) &  0.938 (0.007) &  \textbf{0.984 (0.001)} &  0.943 (0.003) &   0.907 (0.006) &   0.829 (0.01) &  0.947 (0.004) &  0.945 (0.002) &   0.990 (0.002) \\
Supervised         &  0.939 (0.004) &   0.864 (0.019) &  0.812 (0.015) &  0.905 (0.013) &  0.943 (0.004) &  0.977 (0.005) &   0.960 (0.002) &   0.904 (0.019) &  0.832 (0.008) &  0.943 (0.004) &  0.955 (0.001) &  \textbf{0.993 (0.004)} \\
VAT                &  0.949 (0.002) &   0.898 (0.011) &  0.827 (0.017) &  0.929 (0.006) &  0.954 (0.004) &   0.98 (0.006) &   0.960 (0.003) &   0.907 (0.022) &  0.833 (0.005) &  0.952 (0.012) &    \textbf{0.961 (0.001)} &  0.989 (0.002) \\
\bottomrule
\end{tabular}
}
\vspace{.5cm}

\resizebox{0.55\columnwidth}{!}{
\begin{tabular}{l | llllll}
\toprule
Number of labels & \multicolumn{6}{l}{1000} \\
Dataset &           Crop & Electric Devices &          FordB &         Pamap2 &           SITS &          WISDM \\
\hline
Model              &                &                 &                &                &                &                \\
\midrule

Ladder             &   0.78 (0.011) &   0.858 (0.009) &  0.814 (0.008) &  0.899 (0.017) &  0.932 (0.001) &  0.996 (0.001) \\
Logistic Regression &  0.937 (0.001) &    0.91 (0.003) &  0.474 (0.011) &  0.926 (0.004) &  0.934 (0.009) &   0.950 (0.003) \\
Mean Teacher        &  0.966 (0.001) &   0.887 (0.013) &   0.82 (0.006) &   0.96 (0.002) &   0.96 (0.002) &  0.989 (0.005) \\
MixMatch           &   \textbf{0.970 (0.001)} &   \textbf{0.933 (0.003)} &   \textbf{0.859 (0.010)} &  \textbf{0.967 (0.004)} &  0.961 (0.002) &  0.994 (0.001) \\
Random Forest       &  0.959 (0.001) &   0.932 (0.004) &  0.663 (0.021) &  0.948 (0.003) &  0.949 (0.006) &  0.981 (0.001) \\
Self-Supervised     &  0.945 (0.001) &   0.919 (0.007) &   0.828 (0.010) &  0.963 (0.004) &  0.952 (0.006) &    0.994 (0.001) \\
Supervised         &  0.968 (0.001) &    0.910 (0.011) &   0.828 (0.010) &   0.960 (0.003) &   0.960 (0.002) &  \textbf{0.997 (0.001)} \\
VAT                &            0.964 (0.001) &             0.920 (0.008) &            0.828 (0.006) &            0.967 (0.004) &            \textbf{0.965 (0.002)} &            0.995 (0.001) \\

\bottomrule
\end{tabular}
}
\label{tab:performance}
\end{table}

\subsection{Model Ranking}
\begin{table}[h!]
\caption{The average rank of all models based on the wAUC over the $6$ different datasets for various amounts of labels $n_l$. Lower rank indicates stronger model performance. Ranks are shown with decimals due to averaging over datasets.}
\centering
\resizebox{0.3\columnwidth}{!}{
\begin{tabular}{lccccc}
\toprule
& \multicolumn{5}{c}{Number of labels} \\
& 50 & 100 & 250 & 500 & 1000 \\
\hline
\hline
MixMatch           & \textbf{1.7} & \textbf{1.7} & \textbf{1.7} & \textbf{1.7} & \textbf{1.7} \\
VAT                & 2.7 & 2.8 & 2.5 & 2.7 & 2.5 \\
MeanTeacher        & 6.7 & 6.2 & 5.5 & 5.3 & 4.8 \\
Self-supervised    & 4.2 & 4.0 & 3.8 & 4.3 & 4.5 \\
Ladder             & 4.2 & 5.3 & 5.7 & 6.5 & 6.7 \\
Supervised         & 6.3 & 4.7 & 4.8 & 3.7 & 3.2\\
Random Forest      & 3.2 & 3.8 & 4.8 & 4.7 & 5.5 \\
Logistic Regression & 7.2 & 7.5 & 7.2 & 7.2 & 7.2\\
\bottomrule
\end{tabular}
}
\label{tab:ranking}
\end{table}

\subsection{Hyperparameters}
\begin{table}[h!]
\caption{Hyperparameter ranges used for tuning of the different models. Deep Learning models were tuned via Hyperband as described in Section~$3$ while the Random Forest and the Logistic Regression were tuned via Random Search with a bugdet of $100$ model evaluations each.}
\parbox{.5\linewidth}{
\centering
\resizebox{0.25\columnwidth}{!}{
\begin{tabular}{lll}
    \hline
    Parameter & Range & Scale \\
    \hline
    \hline
    \multicolumn{3}{c}{\textbf{Shared}}\\
    \hline
    Weight decay  & [$1e^{-6}$; $1e^{-2}$] & log \\
    Learning rate & [$1e^{-5}$; $1e^{-2}$] & log \\
    Rampup length & [5000; 25000] & linear \\
    Magnitude (RandAug) & [1; 10] & linear \\
    N (RandAug) & [1; 6] & linear \\
    \hline
    \multicolumn{3}{c}{\textbf{VAT}}\\
    \hline
    $\epsilon$ & [0.1; 10.0] & linear \\
    $\alpha$ & [0.1; 5.0] & linear \\
    \hline
    \multicolumn{3}{c}{\textbf{MixMatch}}\\
    \hline
    $\alpha$ & [0.5; 1.0] & linear \\
    $\lambda_{u}$ & [0.0; 150.0] & linear \\
    \hline
\end{tabular}
}
}
\hfill
\parbox{.5\linewidth}{
\centering
\resizebox{0.25\columnwidth}{!}{
\begin{tabular}{lll}
    \hline
    Parameter & Range & Scale \\
    \hline
    \hline
    \multicolumn{3}{c}{\textbf{Self-Supervised Learning}}\\
    \hline
    $\lambda$ & [0.1; 10] & log\\
    horizon $h$ & [0.1, 0.2, 0.3] & discrete \\
    stride $s$ & [0.05, 0.1, 0.2, 0.3] & discrete \\
    \hline
    \multicolumn{3}{c}{\textbf{Ladder Net}}\\
    \hline
    Noise ratio & [0.1, 0.3, 0.45, 0.6] & discrete\\
    Loss weights & [0.1; 10.0] & log \\
    \hline
    \multicolumn{3}{c}{\textbf{Mean Teacher}}\\
    \hline
    $\alpha_{ema}$ & [0.9; 1.0] & log\\
    $w_{max}$ & [0; 10] & linear \\
    \hline
    \multicolumn{3}{c}{\textbf{Random Forest}}\\
    \hline
    Number of trees  & [$100$; $1000$] & linear \\
    Max. tree depth & [$3$; $25$] & linear \\
    \hline
    \multicolumn{3}{c}{\textbf{Logistic Regression}}\\
    \hline
    Regularization term & [None, $L_1$, $L_2$] & discrete \\
    \hline
\end{tabular}
}
}
\label{tab:hyperparameters}
\end{table}

\end{document}